\documentclass[]{gmaart2022}

\renewcommand{\varProcName}{Proc. 33. Workshop Computational Intelligence}
\renewcommand{\varLocation}{Berlin}
\renewcommand{\varDate}{23.-24.11.2023}

\usepackage[utf8]{inputenc} 
\usepackage[T1]{fontenc} 
 
\usepackage{amsmath,amssymb,amsfonts,mathtools,amsthm} % extended math environments

\usepackage[babel,autostyle]{csquotes}

\usepackage{tabularx}
\newcolumntype{L}[1]{>{\raggedright\arraybackslash}p{#1}} % linksbündig mit Breitenangabe
\newcolumntype{C}[1]{>{\centering\arraybackslash}p{#1}} % zentriert mit Breitenangabe
\newcolumntype{R}[1]{>{\raggedleft\arraybackslash}p{#1}} 
\usepackage{booktabs}

\usepackage{paralist}   

\usepackage[noend]{algpseudocode}

\algrenewcommand\algorithmicrequire{\textbf{Voraussetzung:}}
\algrenewcommand\algorithmicensure{\textbf{Abschlussbedingung:}}

\usepackage{listings} % Alternative environment for algorithms
\usepackage{subcaption}  % To place figures side by side, if you wish to do that in latex... (package "subfigure" is deprecated)

\usepackage{epstopdf}
\usepackage{xcolor}
\selectcolormodel{gray}  % only use gray scale colors

\usepackage{scrhack}
\usepackage{varwidth}
\usepackage{rotating} % Rotate text (e.g. in tables)
\usepackage[defaultlines=2,all]{nowidow}  % no widow lines
\usepackage{import}
\usepackage{placeins}
\usepackage{makecell}


\usepackage{amsmath,amssymb,amsfonts,mathtools,amsthm} % extended math environments

\usepackage{longtable} %long tables multiple pages
\usepackage{multirow}
\usepackage{tikz}
\usepackage{pgfplots}
\usepackage{smartdiagram}
\usetikzlibrary{patterns}
\usetikzlibrary{shapes} %flowchart
\usetikzlibrary{shapes.geometric, arrows,positioning}
\usepackage{smartdiagram}
\usesmartdiagramlibrary{additions} % library to create annotations
\AtBeginDocument{
}
\usepackage{xcolor}
\selectcolormodel{gray}

\usepackage[acronym]{glossaries} % no hyper links
\usepackage{hyphenat}
\usepackage{etoolbox}
\usepackage{printlen}
\usepackage{graphicx}
\usetikzlibrary{calc, positioning, matrix}
\usepackage{standalone}
\usepackage{comment}

\newcommand{\newacr}[4][]{\newacronym[
	sort={\ifthenelse{\isempty{#1}}{#2}{#1}},
	]{#2}{#3}{#4}}
\glsdisablehyper

\robustify{\bfseries}

\begin{document}

%\pagestyle{empty}
%\tableofcontents

\hyphenpenalty=2000

\pagenumbering{roman}
%\tableofcontents  %Ausgabe des Inhaltsverzeichnises
%\cleardoublepage
\setcounter{page}{1}
\pagestyle{scrheadings}
\pagenumbering{arabic}

% At least two lines of text at the top and bottom of a page
\setnowidow[2]
\setnoclub[2]

%% Set title and author information
\renewcommand{\Title}{AI-based~automated~active~learning~for discovery~of hidden~dynamic~processes: A~use~case~in~light~microscopy}

% Multiple Institutions
\renewcommand{\Authors}{
    Nils Friederich\textsuperscript{1,2} Angelo Yamachui Sitcheu\textsuperscript{1}, Oliver~Neumann\textsuperscript{1}, Süheyla Eroğlu-Kayıkçı\textsuperscript{2}, Roshan Prizak\textsuperscript{2}, Lennart Hilbert\textsuperscript{2,3}, Ralf Mikut\textsuperscript{1}
}
\renewcommand{\Affiliations}{
    \textsuperscript{1}
    Institute for Automation and Applied Informatics\\
    Karlsruhe Institute of Technology\\
    Hermann-von-Helmholtz-Platz 1\\
    76344 Eggenstein-Leopoldshafen\\
    \textsuperscript{2}
    Institute of Biological and Chemical Systems\\
    Karlsruhe Institute of Technology\\
    Hermann-von-Helmholtz-Platz 1\\
    76344 Eggenstein-Leopoldshafen\\
    \textsuperscript{3}
    Zoological Institute\\
    Karlsruhe Institute of Technology\\
    Fritz-Haber-Weg 4\\
    76131 Karlsruhe
}

% Single Institution							 
%			\renewcommand{\Authors}{Max Mustermann, Thomas Mustermann}
%			\renewcommand{\Affiliations}{Institut, Firma\\
%				Anschrift\\
%				E-Mail: \{max.mustermann,thomas.mustermann\}@xxx.de}
							 
%% Information required for the table of contents:
% Do NOT use superscripts in case of multiple institutions
\renewcommand{\AuthorsTOC}{N.~Friederich, A.Y.~Sitcheu, O.~Neumann, S.~Eroğlu-Kayıkçı, R.~Prizak, L.~Hilbert, R.~Mikut} % List (short) author names without superscripts and conjunction "and".
\renewcommand{\AffiliationsTOC}{Karlsruhe Institute of Technology} % Required for the table of contents (only list names of universities / institutions)

%% Choose caption language of captions (english / german)!
%\setLanguageGerman
\setLanguageEnglish
							 
\setupPaper % generate title and toc
% \setupPaper[titletoc] falls der Titel im Inhaltsverzeichnis vom Titel abweichen soll, bsw. erzwungene Zeilenumbrüche im Titel

% Glossary---------------------------------------------------------
\newacronym{ae}{AE}{AutoEncoder}

\newacronym{ai}{AI}{Artificial Intelligence}

\newacronym{al}{AL}{Active Learning}

\newacronym[
  shortplural={CAE},%
  longplural={Conventional AutoEncoder}%
]{cae}{CAE}{Conventional AutoEncoder}

\newacronym[
  shortplural={CNNs},%
  longplural={Convolutional Neural Networks}%
]{cnn}{CNN}{Convolutional Neural Network}

\newacronym{cpn}{CPN}{Contour Proposal Networks}

\newacronym{cpu}{CPU}{Central Processing Unit}

\newacronym{cv}{CV}{Computer Vision}

\newacronym{db}{DB}{DataBase}

\newacronym{devops}{DevOps}{Development and Operations}

\newacronym{dl}{DL}{Deep Learning}

\newacronym{dna}{DNA}{DeoxyriboNucleic Acid}

\newacronym{e2e}{E2E}{End-to-end}

\newacronym{eapdp}{EAPDP}{Experiment Automation Pipeline for Dynamic Processes}

\newacronym[
  shortplural={EDPs},%
  longplural={Encoded Dynamic Processes}%
]{edp}{EDP}{Encoded Dynamic Process}

\newacronym[
    shortplural={GPUs},%
    longplural={Graphics Processing Units}%
]{gpu}{GPU}{Graphics Processing Unit}

\newacronym{haicore}{HAICORE}{Helmholtz AI COmpute REssources}

\newacronym{hac}{HAC}{Hierarchical Agglomerative Clustering}

\newacronym{kaida}{KaIDA}{Karlsruhe Image Data Annotation}

\newacronym{lammps}{LAMMPS}{Large-scale Atomic/Molecular Massively Parallel Simulator}

\newacronym{mae}{MAE}{Masked AutoEncoder}

\newacronym{ml}{ML}{Machine Learning}

\newacronym[
  shortplural={MLOps},%
  longplural={Machine Learning Operations}%
]{mlop}{MLOp}{Machine Learning Operation}

\newacronym{mse}{MSE}{Mean Squared Error}

\newacronym{mst}{MST}{Minimum Spanning Tree}

\newacronym{nacip}{NACIP}{Natural, Artificial and Cognitive Information Processing}

\newacronym{pca}{PCA}{Principal Component Analysis}

\newacronym{poc}{PoC}{Proof of Concept}

\newacronym{pol2}{Pol II}{Polymerase II}

\newacronym{rhel}{RHEL}{Red Hat Enterprise Linux}

\newacronym{rna}{RNA}{RiboNucleic Acid}

\newacronym[%
  shortplural={RoIs},%
  longplural={Regions of Interest}%
] {roi}{RoI}{Region of Interest}

\newacronym{ssl}{SSL}{Self-Supervised Learning}

\newacronym{sota}{SOTA}{State-Of-The-Art}

\newacronym{ui}{UI}{User Interface}

\newacronym[
  shortplural={VAE},%
  longplural={Variational AutoEncoder}%
]{vae}{VAE}{Variational AutoEncoder}

\newacronym{xai}{XAI}{Explainable Artificial Intelligence}
%--------------------------------------------------------------------

%% Insert content here:

\section*{Abstract}
\label{sec:abstract}

In the biomedical environment, experiments assessing dynamic processes are primarily performed by a human acquisition supervisor. Contemporary implementations of such experiments frequently aim to acquire a maximum number of relevant events from sometimes several hundred parallel, non-synchronous processes. Since in some high-throughput experiments, only one or a few instances of a given process can be observed simultaneously, a strategy for planning and executing an efficient acquisition paradigm is essential. To address this problem, we present two new methods in this paper. The first method, \gls{edp}, is \gls{ai}-based and represents dynamic processes so as to allow prediction of pseudo-time values from single still images. Second, with \gls{eapdp}, we present a \glspl{mlop}-based pipeline that uses the extracted knowledge from \gls{edp} to efficiently schedule acquisition in biomedical experiments for dynamic processes in practice. In a first experiment, we show that the pre-trained \gls{sota} object segmentation method \gls{cpn} works reliably as a module of \gls{eapdp} to extract the relevant object for \gls{edp} from the acquired three-dimensional image stack.

%--------------------------------------------------------------------
%--------------------------------------------------------------------
%--------------------------------------------------------------------

\section{Introduction}
\label{sec:introduction}

For the imaging-based assessment of dynamic processes in biomedical settings, objects of interest must be identified and relevant events that characterize the dynamic process must be recorded during their time of occurrence. Commonly, a human operator controls the imaging instrument to examine a biomedical sample using a microscope and relevant objects are found in the sample by inspection. Alternatively, the operator estimates for each object of interest the time at which an event of interest is expected to occur based on previous experience and triggers the recording of the event at that time. Nevertheless, many contemporary experiments provide several hundred relevant objects that can, in principle, be imaged in parallel. Events of interest, however, are non-synchronous and the estimation of future event times requires extensive human effort, is prone to error, and not necessarily time-efficient. These obstacles can result in unnecessarily large amounts of irrelevant data, unnecessary experimental repeats, or experimental biases inflicted by additional light exposure of the sample~\cite{phototoxicity}. To address these obstacles, we present two new methods for the automated, real-time planning and execution of such experiments.

\paragraph{\gls{edp}.} The traditional method of capturing all data or relying on human experience to predict future events is outdated and inefficient. Instead of relying on human experience, an accurate and comprehensive model of the dynamic process should be created. This model should be capable of uniquely identifying a relevant object state through a process known as fingerprinting, similar to how humans do it. In biomedicine, it is crucial that this fingerprint remains consistent despite contextual changes such as noise, brightness changes, or affine transformations. By modeling the relationship between the fingerprints, the dynamic process can be represented orderly. This representation suits as an approximation of relative progress within the dynamic process. This relative progress can also be interpreted as a relative time, known as pseudo-time. By adopting this method, we can achieve efficiency in predicting future events and a deeper understanding of the dynamic process.

\paragraph{\gls{eapdp}} To unlock the full potential of the \gls{edp}, a well-designed pipeline is an absolute must. Such a pipeline should be able to recognize specific states in the real world, identify relevant objects, and then calculate a pseudo-time for those states using the \gls{edp}. Armed with this knowledge, the pipeline can automatically plan and execute a new state capture for any significant event that occurs. Due to the uncertain nature of the \gls{edp} predictions, the pipeline must be able to respond to unsuccessful recordings and learn from them. This is where \glspl{mlop}~\cite{what_mlops_is} comes in. By retraining an existing production model in accordance with the live context, \glspl{mlop} ensures that the pipeline always uses a current and accurate model, resulting in a potentially better outcome in real-world experiments.

%--------------------------------------------------------------------
%--------------------------------------------------------------------
%--------------------------------------------------------------------

\section{Related Work}
\label{sec:related_work}

\paragraph{Object Extraction.}
Basically, there are different possibilities in \gls{cv} like object detection and object segmentation, to identify individual objects in an image~\cite{cv_algos_hw_impl_survey} and thus extract them. In the biomedical context, many methods mostly focus on segmentation~\cite{cell_segmentation_common} with \gls{sota} methods like StarDist~\cite{stardist} and \gls{cpn}~\cite{celldetection}.

\paragraph{Pseudo-time predictions.}
A first approach for pseudo-time predictions with classical, non \gls{dl} methods was presented in~\cite{hilbertlab_pseudotime}. For extracting relevant objects from the acquired image, thresholding is used as a classical \gls{cv} segmentation method. Then the object's fingerprint is generated linearly with a \gls{pca}~\cite{pca_jackson,pca_ian}. However, biological processes are usually not linear~\cite{nonlinear_biology}. Therefore, recently, non-linear encodings of the dynamic processes using \gls{dl} methods have become popular~\cite{dl_pseudotime_alzheimer,reconstruction_fabian,cell_cyclus_stegmaier,rappez2020deepcycle}. For example, \cite{cell_cyclus_stegmaier} and \cite{reconstruction_fabian} present \gls{dl} approaches to encode cell cycles and derive predefined cell phases. However, this classification-based approach does not allow for deriving continuous relations like the pseudo-time to each other directly. This continuous relation was modeled with DeepCycle in \cite{rappez2020deepcycle}. The training of DeepCycle is performed supervised. For this purpose, virtual labels are calculated based on the fluorescence intensity in specifically labeled channels of the cells. These classes can then be used as anchor points during training to determine a (relative) cell state as a pseudotime. It is important to note that the assumption that a correlation of fluorescence intensity to cell phase can be used is not always true in the biological context. A \gls{dl} method that follows a comparable pseudo-time approach to the given constraints, in this paper, was presented in~\cite{dl_pseudotime_alzheimer}. In~\cite{dl_pseudotime_alzheimer}, an \gls{ae} approach for pseudo-time approximation is used as a \gls{ssl} approach. A \gls{dl} model is used as the \gls{vae}~\cite{vae_first} encoding, from whose \gls{hac} and \gls{mst} code the pseudo-time is then determined. However, this approach also has a few limitations. First, a recording necessarily contains exactly one relevant object in one acquired image. Second, the entire dataset was acquired under comparable acquisition conditions, which also only contain identical positioned and oriented objects and are not able to learn affine transformations~\cite{affine_invariance_revisited} between objects. Both constraints are generally not satisfied for microscopic images, such as in~\cite{microscope_Data_pol2_klf2b,microbeseg_data}. Furthermore, this pseudo-time method was not designed as an \gls{e2e} model, which deprives the \gls{dl} model of the ability to internally bind affine transformed objects.

\paragraph{AutoEncoder.} 
Autoencoders are \gls{ssl} methods to learn a representation from a given suitability, such as an image~\cite{mae_survey,convenxtv2}. For example, the autoencoder can be represented by a \gls{cae}~\cite{mae_survey} and/or a \gls{vae}~\cite{vae_first}. Especially recently, \glspl{mae}~\cite{transformer_mae} have become more popular than \gls{cae} because of their ability for a better visual representation learning~\cite{mae_survey}, either using a transformer-based approach \cite{transformer_mae} or the \gls{cnn}-based approach~\cite{convenxtv2}. However, since in~\cite{convenxtv2} the higher efficiency of ConvNeXt V2 is shown, this model is chosen for this work. In addition to the pure learning of a visual representation in the form of a fingerprint, relations between the fingerprints can also be learned, e.g. with \gls{vae}~\cite{dl_pseudotime_alzheimer}. 

\paragraph{\gls{ssl}}
To train the \gls{dl} model in a supervised manner, labeled data are generally rare in the biomedical domain~\cite{ssl_data_effizient_learning}. There are \glspl{db} like BioMedImage.io~\cite{bioimageio} or several challenges with own datasets~\cite{grand_ch_1,grand_ch_2,cell_tracking_challenge}. However, especially for biological datasets with sometimes hundreds of relevant objects in an image, the datasets are often limited to the 2D case. Furthermore, in the context of this work, labeling the relevant events with pseudo-time stamps is only approximate, demanding, error-prone and time-consuming. Therefore, unsupervised or \gls{ssl} methods are often used in the biomedical context~\cite{3d_bioimage_time_consuming_labeling}. Thereby, \gls{al}~\cite{al_survey_08} is used to selectively integrate expert knowledge into the learning process. Since \gls{al} aims to keep the number of interactions to a minimum~\cite{al_disadvatage,al_survey_21}, data-efficient learning is preferable. For example, existing datasets from a related context can be identified and leveraged to train more robust models in transfer learning~\cite{figerprint_lena_maier_heine,fingerprint_marcel,angelo_ci_2023}. In order to be able to use possibly directly existing pre-trained models from similar contexts, a new concept was developed in~\cite{angelo_ci_2023}.

%\paragraph{\glspl{mlop}}
%\glspl{mlop} is seen as the intersection of machine learning, DevOps, and data engineering to increase automation and improve the quality of production models~\cite{what_mlops_is}. 

%--------------------------------------------------------------------
%--------------------------------------------------------------------
%--------------------------------------------------------------------

\section{Methodology}
\label{sec:methodology}

\subsection{\gls{edp}}
\label{subsec:Methodology-EDP}
\begin{figure}[tb]
        \centering
        \includegraphics[width=0.75\linewidth]{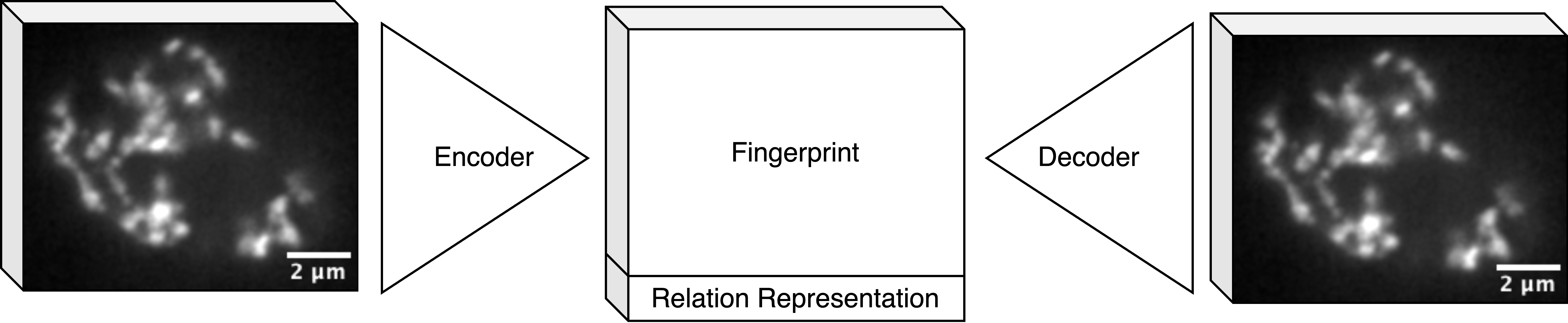}
        \caption{Visualization of the \gls{edp} model. A 3D object is transformed by an encoder into a fingerprint (like \gls{mae}) and into a relation representation between the fingerprints (like \gls{vae}). The example object is a recording of the DNA channel of a nucleus from an internal zebrafish embryo dataset. A scale bar of $2 \mu m$ is indicated at the bottom of the input/output image.}
        \label{fig:edp_method_idea}
    \end{figure}

\begin{figure}[tb]
        \centering
        \includegraphics[width=0.95\linewidth]{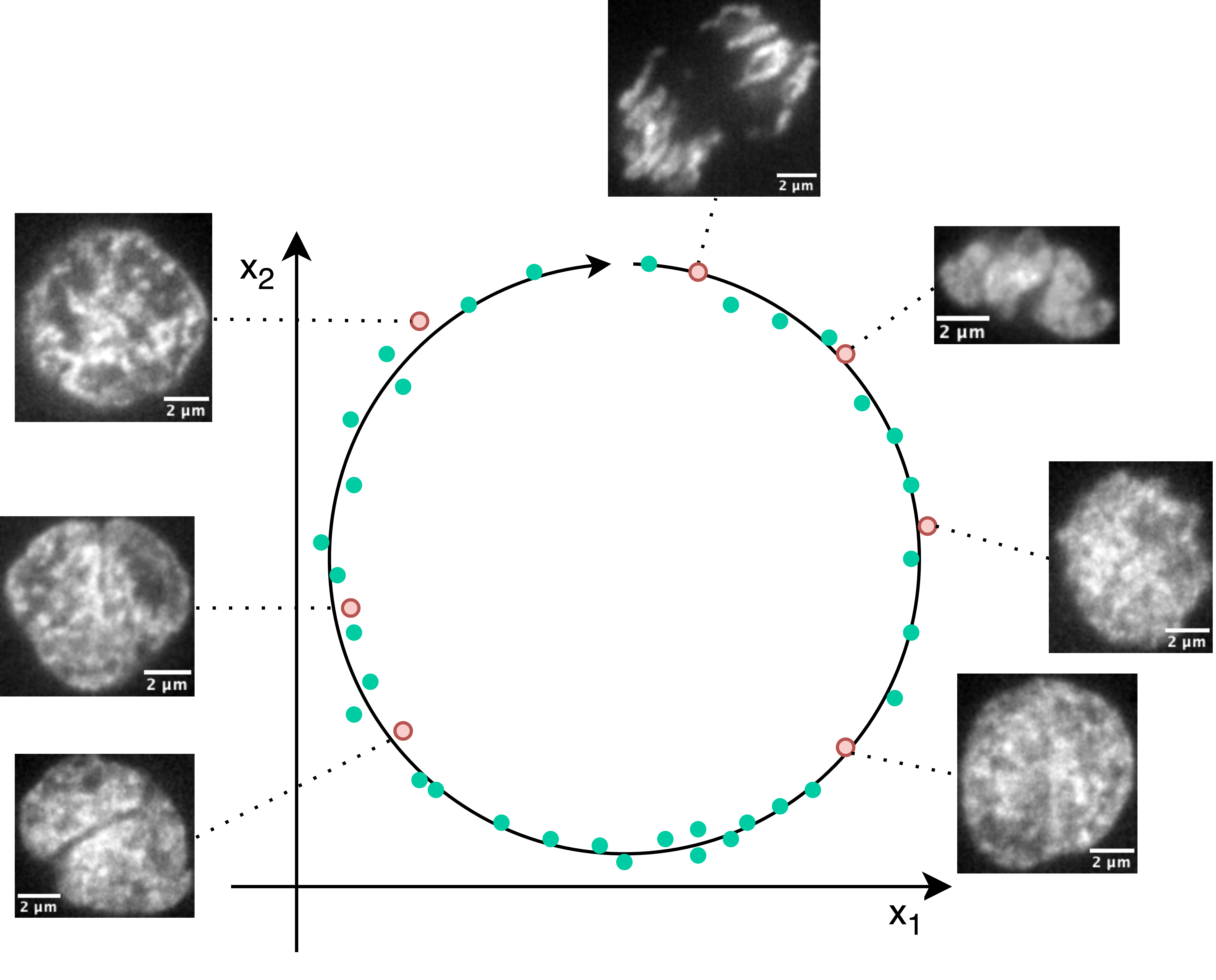}
        \caption{Example of a 2D feature space representation for an encoded dynamic process. Each point represents an encoded image. The circle serves as an estimation for the positioning of points in its vicinity. It's worth noting that due to the presence of uncertainty, the points may not always be precisely on the circle, but rather in its proximity. For the seven red dots, example images of cell nuclei from zebrafish embryos at various stages of cell division are shown. A scale bar of $2~\mu m$ is indicated at the bottom of each example image. The images are from an internal dataset.}
        \label{fig:pseudotime_vae_example}
    \end{figure}

For the modeling of the dynamic process, a new concept is introduced with EDP. The new concept of EDP is based on an \gls{ae} and is visualized in Figure~\ref{fig:edp_method_idea}.
The basic idea of the new concept is to separate the generation of the fingerprint from the learning of the relation as a representation between all states. The fingerprint generation is done using a \gls{mae} as an evolution of \gls{cae}. Specifically, the \gls{sota} \gls{mae}-based method ConvNeXt V2 is chosen. During the learning process, a maximum recovery of the encoded image is aimed in accordance with \gls{ssl}. According to the challenge posed by biological objects, a context-independent representation is required. For this purpose, the images can be modified during the learning process through Data~Augmentation~\cite{data_aug_regularization} 
techniques like Rotations, reflections, contrast adjustments or noise additions.\\
In addition to the fingerprint, the relation must also be learned as the actual modeling of the dynamic process. For this purpose, a \gls{vae}-like modeling is used by learning the uncertainty $v$ in addition to the circle angle $\alpha$. The assumption is that objects succeeded each other in the dynamic process with the relative distance $t$ corresponding to this relative distance and differ in the same ratio in the circle representation. Such an exemplary circle representation is shown in Figure~\ref{fig:pseudotime_vae_example} using a cell division process of the zebrafish embryo. The state of the cell after cell division is visualized at 00 o'clock and up to the state of the cell just before cell division at 11 o'clock. This corresponds to a relative distance of $\sim$0.92 (normalized between [0,1)). This temporal difference must also be valid in reality for the temporal distance according to the model statement.

\subsection{\gls{eapdp}}
\label{subsec:methodology_mlops}
\begin{figure}[tb]
        \centering
        \includegraphics[]{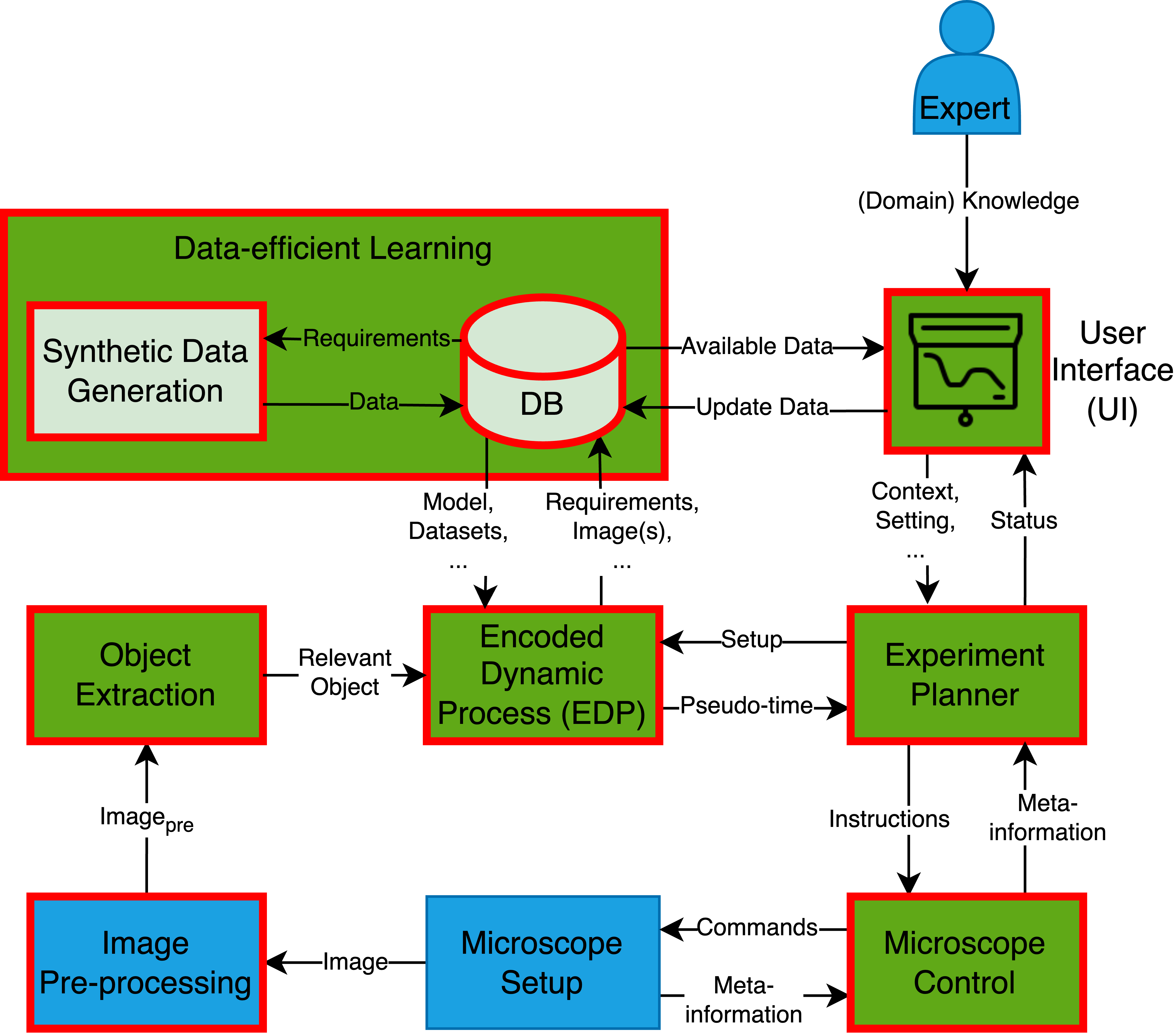}
        \caption{\glspl{mlop} pipeline with the new \gls{edp} module. \gls{ai}-based \glspl{mlop} modules are marked with a green background, non \gls{ai}-based ones with a blue background. Additionally, all modules marked with a red border must be newly developed or only partially adapted from existing methods.
        }
        \label{fig:mlops_pipeline_diagram}
    \end{figure}

The new \gls{edp} module is integrated as a module into the new \glspl{mlop}-based pipeline \gls{eapdp}. The pipeline concept is visualized in Figure~\ref{fig:mlops_pipeline_diagram} and contains nine other modules besides the \gls{edp} module. Each of these ten modules is briefly described below. The explanation of the modules and their relationships to each other is based on the pipeline visualization of Figure~\ref{fig:mlops_pipeline_diagram}.

\paragraph{Microscope setup.} In the \gls{eapdp}, the microscope is used as an actuator to the real-world environment represented by a biomedical sample. For this purpose, all microscope components relevant to image acquisition and the microscope accessories, such as lasers in the case of a laser scanning microscope, must be controllable via appropriate interfaces of the specific microscope setup. In addition, the microscope must be able to react on given commands like image acquisition or requested meta-information like the objective position in a standardized way.

\paragraph{Image Pre-processing.} To optimize the analysis of dynamic processes in the biomedical environment, the raw images acquired through experiments must be pre-processed according to the microscope setup and the context of the targeted event. This may involve methods such as cropping, contrast adjustment, or denoising. Various libraries, such as Albumentations~\cite{albumentations}, offer pre-processing methods that can be used to improve the quality of the images and optimize their analysis by other modules in the machine learning operation pipeline.

\paragraph{Object Extraction.} During an acquisition, the relevant object and the surrounding context are captured. In order to better analyze the object, it is necessary to separate it from the surrounding context. The extraction from the whole image is done via pre-trained segmentation methods. To find a suitable method, we compare the \gls{sota} cell segmentation algorithms StarDist and \gls{cpn} using a microscopic dataset in a first experiment. Importantly, the actual pseudo-time determination cannot be performed if both methods' segmentation is insufficient. Therefore, this submodule is of particular importance. Because (well) labeled data are generally scarce in the biological context, this work evaluates generalization performance during inference with already pre-trained models on new, unknown images. Since there are only pre-trained weights for 2D segmentation for both methods, the dataset was split into 2D images along the z-axis.

\paragraph{\gls{edp}.} The \gls{edp} module gets the extracted object and should pass the pseudo-time to the experiment planner. In order to do this, the module is equipped by the Experiment planner before with the appropriate experiment setup. With the setup, the \gls{edp} model can then query according to its existing knowledge like pre-trained models in the context of data-efficient learning. If no weights are available, training can also be done with/without \gls{al} as specified by the expert. After successful training, the model is passed to the \gls{db} with the appropriate required metadata for possible further use. Then, when the inference with the original extracted relevant object has been determined, the results are passed to the expert planner accordingly. The recorded inference image is also sent to the Data-efficient Learning module and stored in its \gls{db}.

\paragraph{Experiment planner.} The Experiment planner is the central module of the experiment automatization. As input it gets the pseudo-times for the recorded objects. Based on the experiment context, including interesting events, the Experiment planner can plan future experiments with utmost precision. Once the plan is set, the Experiment planner gives the microscope the command to ensure that the image captures the object's state at the right time, leaving no room for errors. Additionally, it can query the state of the microscope to ensure that there was no hardware drift, such as when moving to the object position. All the information about the experiment's state is then passed on to the \gls{ui}, ensuring that all aspects of the experiment are under control.

\paragraph{\gls{ui}.} The \gls{ui} is the interface between the expert and the \glspl{mlop} pipeline. On the one hand, simple interactions can be provided, such as displaying meta-information or adapting the experimental context, e.g. the cell classes that occur. On the other hand, much more complex interactions such as result justifications of \gls{dl} models can be represented through \gls{xai}~\cite{xai_survey} or expert knowledge can be brought into the pipeline within the context of \gls{al}. With \gls{xai}, the expert should be able to understand better the processes in the \gls{dl} models used and why decisions were made, e.g. for event detection. This helps the expert to eliminate potential errors like unfavorable experiment settings at an early stage. Such \gls{xai} methods can be realized using a library like PyTorch Captum~\cite{captum}. For \gls{al}, only if the expert can capture the actual state in the best possible way, the expert can transfer his domain knowledge to the method in the best possible way and support the method. For example, points, boxes, or entire segmented regions can be passed to the method as hints. For this purpose, a custom segmentation module can be developed based on exiting \gls{al} labeling platforms like ObiWanMicrobi~\cite{obiwanMicroby_microbseg} or \gls{kaida}~\cite{kaida}.

\paragraph{Expert (Domain) Knowledge.} The domain knowledge contributed by the expert to the \glspl{mlop} pipeline can take several forms. For example, the context of the experiment with a specific cell class can improve a more efficient event detection module. Furthermore, knowledge can be injected, e.g. by labeling in the context of \gls{al}. For this, the expert must ensure the quality of the injected domain knowledge with maximum correctness. Incorrect information can affect the learning processes in the network.

\paragraph{Data-efficient Learning.} To minimize \gls{al} interactions with the human expert, as much existing knowledge as possible is reused. To this end, building on \cite{angelo_ci_2023}, a new \gls{ae}-based fingerprinting approach for datasets and \gls{ml} models is being implemented to reuse as much knowledge as possible. For this, from a \gls{db}, context-given requirements can query existing knowledge. If no data is available, it can be created synthetically e.g. with biophysical simulations like \gls{lammps}~\cite{LAMMPS}.

\paragraph{Microscope control.}
In order for the planned experiment to be automated and performed in real-time, a corresponding software library is needed to control the microscope. Since the first release in 2010, $\mu$Manager has been used for this purpose as one of the \gls{sota} open-source solutions~\cite{micropilot,high_throu_microfluidic,mmagellan,autoscanj}. Therefore, this is also used in this work.

\subsection{Exemplary Use Cases}
\label{subsec:use_cases}
Example use cases for the presented \gls{eapdp} with the \gls{edp} are presented using record extracts in Figure~\ref{fig:intro} below.
A first use case is shown in Figure~\ref{fig:biomed_ds_1} and represents the temporal sorting of \gls{rna} \gls{pol2} clusters that occur in the nuclei of pluripotent zebrafish embryos. A method for this use case has already been presented in \cite{hilbertlab_pseudotime}. A comparison of the pipeline based on classical \gls{ml} methods with our \gls{dl}-based \gls{edp} method allows a direct statement about limitations or improvements of our approach. 
The second use-case in Figure~\ref{fig:biomed_ds_2} is the recording of cell divisions in pluripotent zebrafish embryos, where the time of reaching a new division stage and thus the regions of an event of interest need to be extrapolated. A final biological application from the field of microbiology is presented in Figure \ref{fig:biomed_ds_4}. In this example, one interesting event could be the state at which $n$ microbes reach the recording region. For this purpose, a modeling of the cell division process with \gls{edp} can be used to plan the experiment accordingly and automatically record the event of interest at a time $t$. The modeling of the cell division process with \gls{edp} can be used for this purpose.

    \begin{figure}[tb]
        \centering
        \begin{subfigure}[b]{0.3\textwidth}
            \centering
            \includegraphics[height=2.25cm]{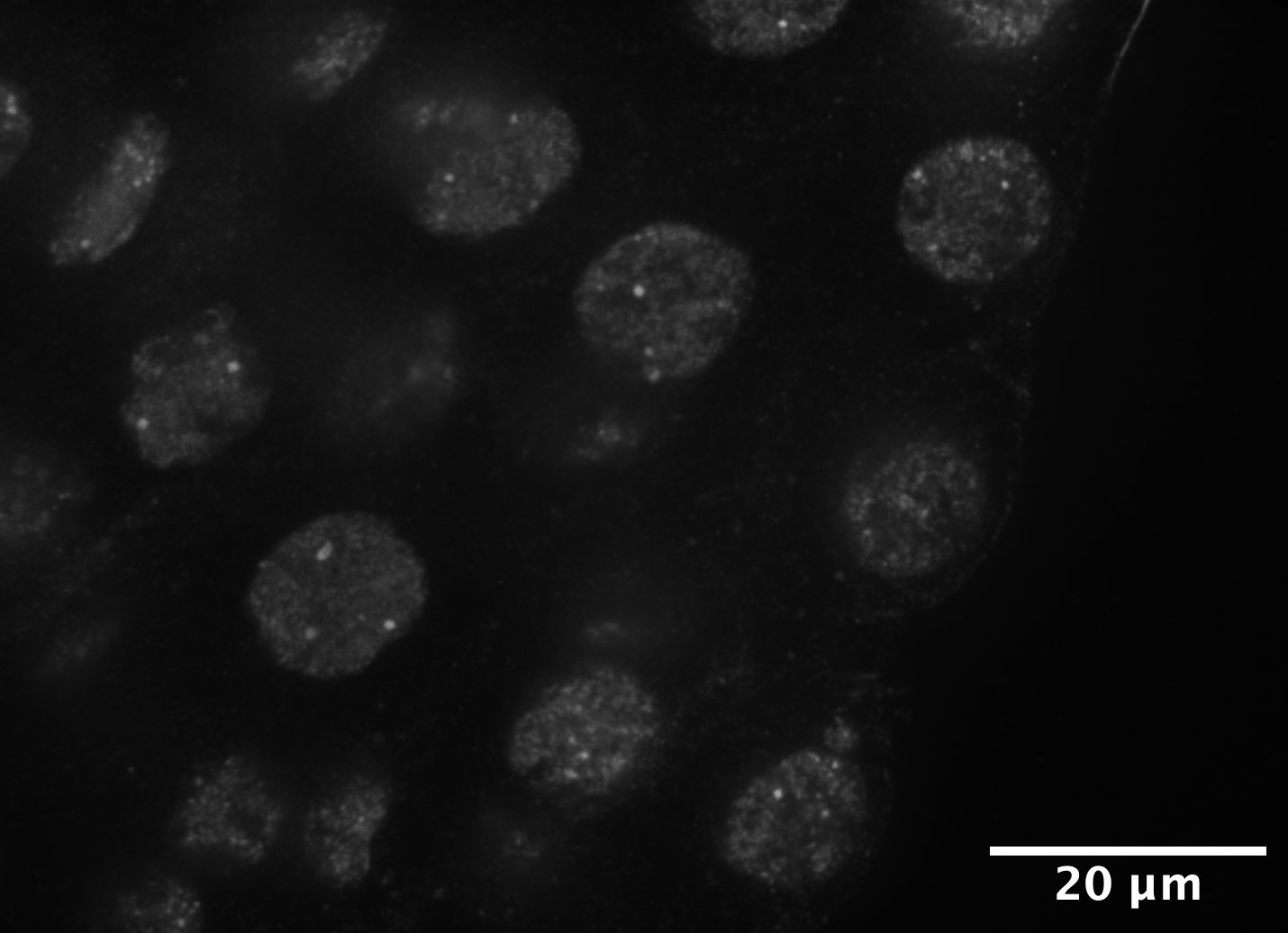}
            \caption{\cite{microscope_Data_pol2_klf2b}}
            \label{fig:biomed_ds_1}
        \end{subfigure}
        \begin{subfigure}[b]{0.3\textwidth}
            \centering
            \includegraphics[height=2.25cm]{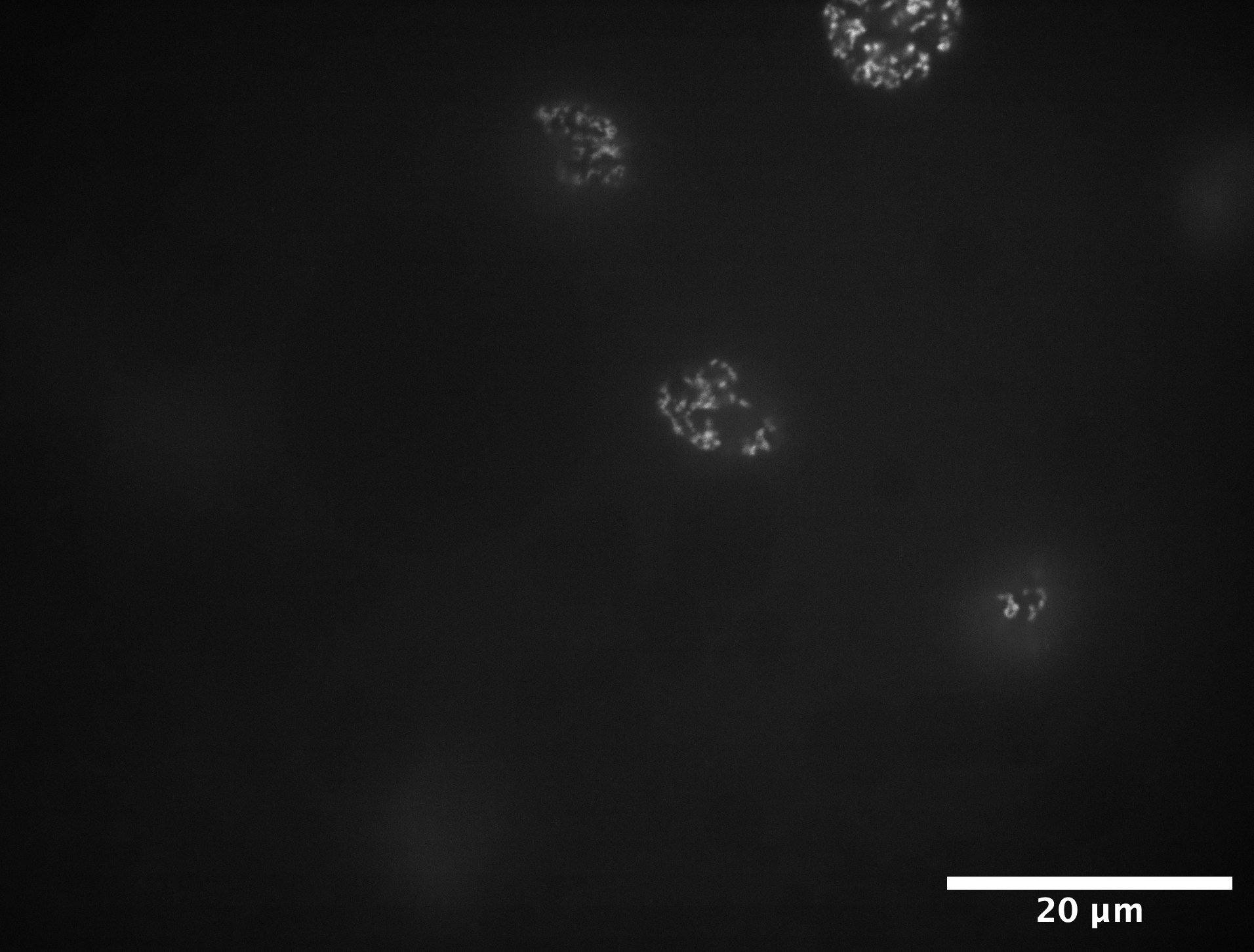}
            \caption{\textit{Internal dataset}}
            \label{fig:biomed_ds_2}
        \end{subfigure}
        \begin{subfigure}[b]{0.3\textwidth}
            \centering
            \includegraphics[height=2.25cm]{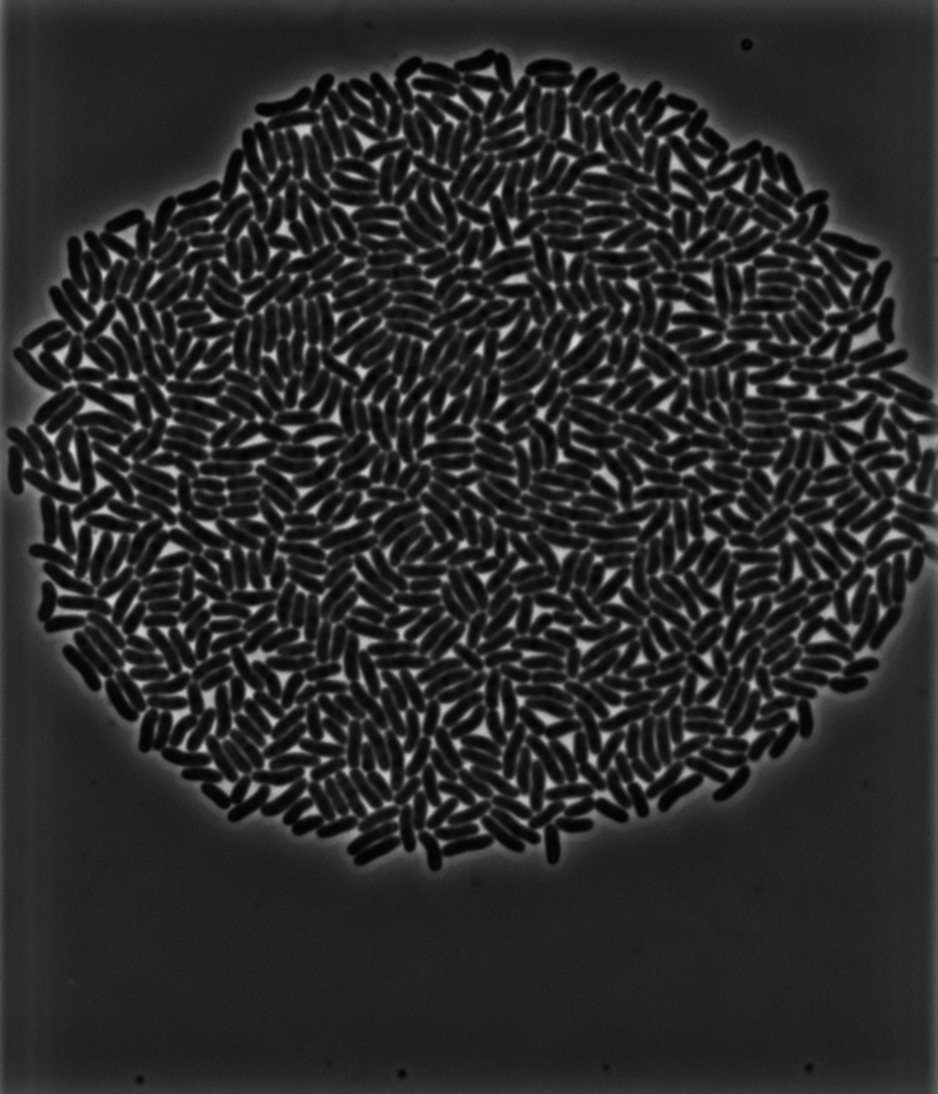}
            \caption{\cite{data_microbs_tracking_million}}
            \label{fig:biomed_ds_4}
        \end{subfigure}
    
        \caption{Three exemplary images from biological datasets for dynamic processes. Figure~\ref{fig:biomed_ds_1} shows cell nuclei of zebrafish embryos with marked Pol II Ser5P clusters. Figure~\ref{fig:biomed_ds_2} shows the \gls{dna} of zebrafish embryos nuclei. In these first two images, a scale bar of $20 \mu m$ is shown in the lower left. The last Figure~\ref{fig:biomed_ds_4} shows a microbial cell division state.}
        \label{fig:intro}
    \end{figure}

In addition to these biological use cases, other use cases are also possible, e.g. in medicine. For example, by modeling a tumor accordingly, a prediction can be made about the relative stage. Consequently, a therapy concept such as surgery or medication can be tailored to the patient.

%--------------------------------------------------------------------
%--------------------------------------------------------------------
%--------------------------------------------------------------------

\section{Experiments}
\label{sec:Experiments}
The comparison of segmentation algorithms was performed on \gls{haicore} resources equipped with Intel Xeon Platinum 8368 \glspl{cpu} and an Nvidia A100-40 \gls{gpu}~\cite{kit_haicore_hw_overview}. The operating system utilized was \gls{rhel} version 8.6.

\subsection{Dataset}
\label{subsec:experiments_dataset}
The internal microscope dataset from Figure~\ref{fig:biomed_ds_2} is used to compare the segmentation algorithms. This dataset was chosen over the other two example datasets from Figures~\ref{fig:biomed_ds_1} and~\ref{fig:biomed_ds_4} because of the challenging, frayed structure of the nuclei as the relevant image object. This is because the fibrillar structure of the nuclei sometimes deviates strongly from their typical ellipsoidal shape as in Figure~\ref{fig:biomed_ds_1} due to individually advanced cytokinesis. This poses a challenge because contiguous pixel regions are not trivially identifiable and correct boundary segmentation is a challenge. With microbeSEG~\cite{microbeseg}, a working \gls{sota} solution for microbes like in Figure~\ref{fig:biomed_ds_4} also already exists.

For the dataset in Figure~\ref{fig:biomed_ds_2}, zebrafish embryo DNA was imaged. DNA was stained with 1:10000 5'-TMR Hoechst in TDE or glycerol. Confocal z-sections were obtained using a commercial instant SIM microscope (iSIM, VisiTech). A Nikon 100x oil immersion objective (NA 1.49, SR HO Apo TIRF 100xAC Oil) and a Hamamatsu ORCA-Quest camera were used for image acquisition. In accordance with a common problem in biology, no labels exist for this dataset. According to the desired 2D segmentation, the 3D images are split into 2D images along the z-axis.

\subsection{Object extraction}
\label{subsec:Experiments_Segmentation} 
\begin{figure}[tb]
        \centering
        \begin{subfigure}[b]{0.3\textwidth}
            \includegraphics[width=\linewidth]{Images/stardist_original_130.png}
            \caption{Original}
            \label{fig:results_seg_1}
        \end{subfigure}
        \begin{subfigure}[b]{0.3\textwidth}
            \includegraphics[width=\linewidth]{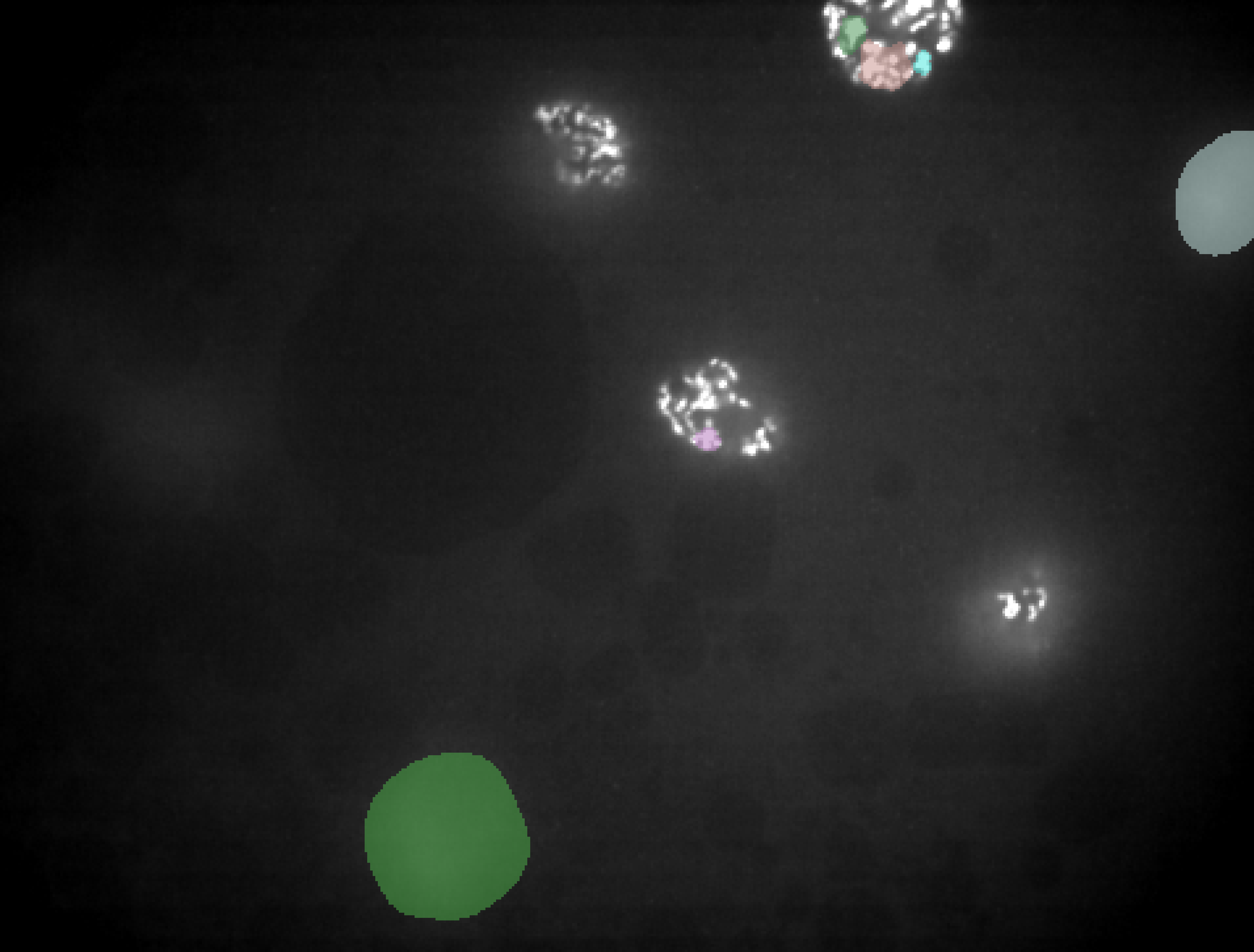}
            \caption{StarDist~\cite{stardist}}
            \label{fig:results_seg_2}
        \end{subfigure}
        \begin{subfigure}[b]{0.3\textwidth}
            \includegraphics[width=\linewidth]{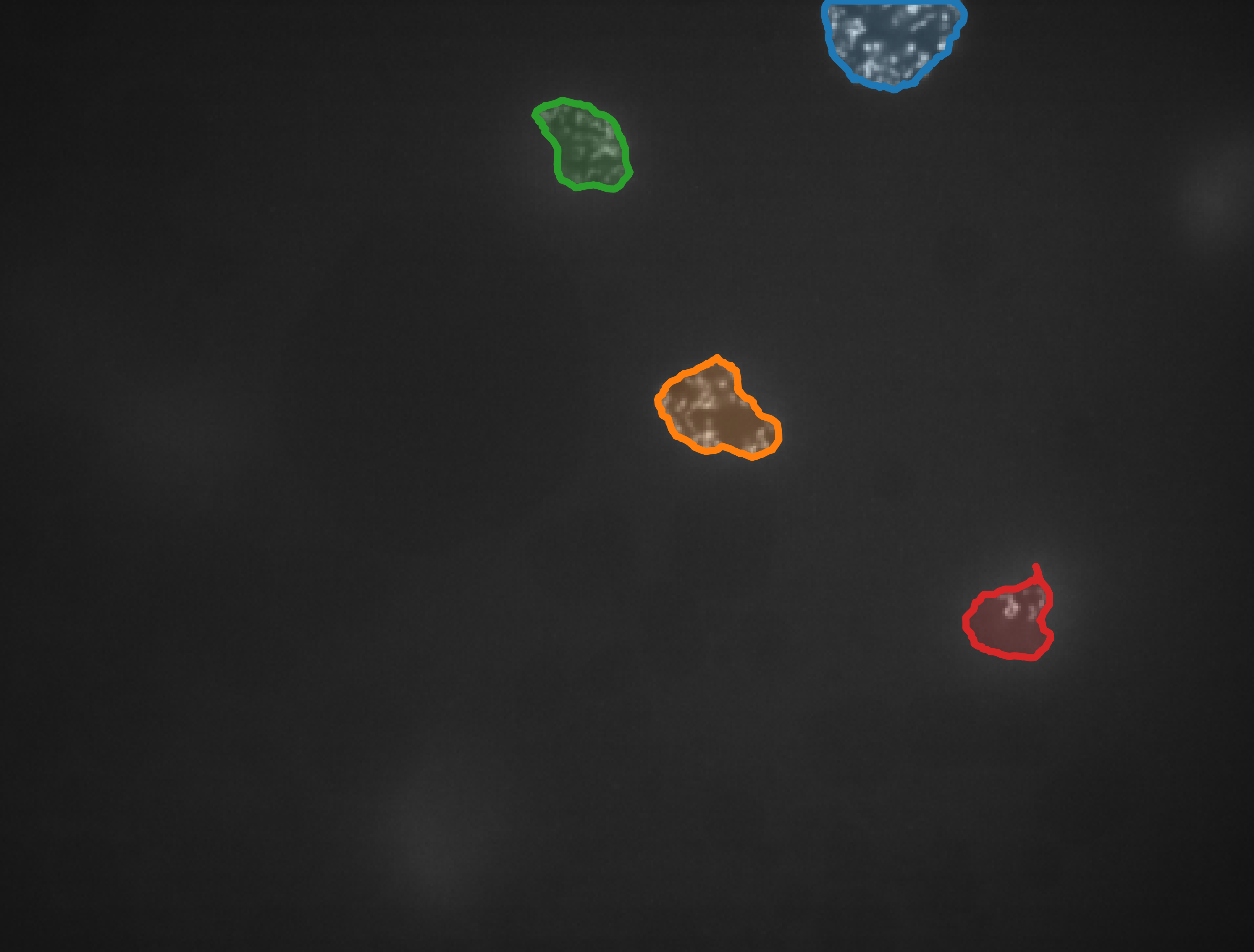}
            \caption{\gls{cpn}~\cite{celldetection}}
            \label{fig:results_seg_3}
        \end{subfigure}
        
        \caption{Comparison of segmentation predictions for the two \gls{sota} methods StarDist~\cite{stardist} and \gls{cpn}~\cite{celldetection}. Figure~\ref{fig:results_seg_1} represents the original image, duplicated from Figure~\ref{fig:biomed_ds_2}. In Figure~\ref{fig:results_seg_2} and Figure~\ref{fig:results_seg_3}, the predictions of pre-trained StarDist or \gls{cpn} are then shown. The predictions are highlighted differently for better visual differentiation depending on the method used.
        }
        \label{fig:results_seg}
    \end{figure}

These 2D images were then segmented using each of the two methods. In the following, the results are evaluated qualitatively because of the non-existent labels. Therefore, the results are shown in Figure~\ref{fig:results_seg}. The comparison of the original image in Figure~\ref{fig:results_seg_1} with the StarDist prediction in Figure~\ref{fig:results_seg_2}, shows that StarDist cannot well segment semantically related objects as the nucleus in the upper area. For the method designs with center prediction, StarDist focused primarily on segmenting ellipsoidal objects from~\cite{dsb2018} and was trained only on these. The cell detection was designed to be more flexible and additionally trained on a more heterogeneous set of non-elliptical cells such as \textit{MCF7} from the dataset~\cite{livecell}. This leads to better generalization and results in qualitatively evaluated good initial segmentation performance on this most challenging of our datasets from Figure~\ref{fig:intro}.

Thus, we could show that \gls{cpn} is a good pre-trained \gls{sota} approach for extracting the relevant objects from the 2D decomposition. The 2D segmentations can be reassembled back to 3D segmentations in post-processing, e.g. using Nearest Neighbor. Based on this, the further submodules of \gls{edp} can be developed in future work and the presented \glspl{mlop} pipeline can be built upon it.

%--------------------------------------------------------------------
%--------------------------------------------------------------------
%--------------------------------------------------------------------

\section{Conclusion and Further Work}
\label{sec:conclusion}
\begin{sloppypar}
%Conclusion
In this work, we motivated that due to the large number of parallel non-synchronous dynamic processes, a novel concept for automated planning and execution of two novel \gls{dl}-based approaches is essential. First, the \gls{edp} was introduced to model dynamic processes and derivate a pseudo-time for a given object state. The pseudo-time prediction can then be used with the \gls{eapdp} for real-time experiment automation. We explained the \gls{edp} realized within the \glspl{mlop} pipeline by an \gls{ae} and trained using \gls{ssl} with \gls{al}. At the same time, the key advantage of higher execution speed and lower human cost while minimizing user interactions with data-efficient learning was highlighted. Finally, as a first \gls{poc}, we showed the necessary pre-processing step for the \gls{edp} to extract the relevant objects based on good inference results of \gls{cpn}.
\end{sloppypar}

%Further work
However, the lack of pre-trained weights for 3D segmentation was a drawback of the segmentation experiments. However, since the fragmented objects are partially reconnected along the z-axis, this could simplify the problem and improve accuracy. This will be done as soon as appropriate weights are available. In addition, a suitable affine-invariant 3D \gls{ae} needs to be developed for use within the \gls{edp} method. In this context, further research is needed to investigate whether the ConvNeXt V2 is suitable for 3D segmentation, also from an efficiency perspective. Of course, the modules of the \glspl{mlop} pipeline must be implemented accordingly.

%--------------------------------------------------------------------
%--------------------------------------------------------------------
%--------------------------------------------------------------------

\section*{Acknowledgments}

This project is funded by the Helmholtz Association under the Program \gls{nacip} and the Helmholtz Association’s Initiative \& Networking Fund through Helmholtz AI.  All experiments were performed on the \gls{haicore}. We sincerely thank all of them for supporting our research.

The authors have accepted responsibility for the entire content of this manuscript
and approved its submission. We describe the individual contributions of
N. Friederich (NF), A. Yamachui Sitcheu (AYS),  O. Neumann (ON), S. Eroglu-Kayıkçı (SEK), R. Prizak (RP), L. Hilbert (LH), R. Mikut (RM): Conceptualization: NF, LH, RM; Methodology: NF, LH, RM; Software: NF; Investigation: NF; Resources: SEK, RP, LH; Writing – Original Draft: NF; Writing – Review \& Editing: NF, AYS, ON, SEK, RP, LH, RM; Supervision: LH, RM; Project administration: RM; Funding Acquisition: LH, RM

%--------------------------------------------------------------------
%--------------------------------------------------------------------
%--------------------------------------------------------------------

%\clearpage
%% Bibliography
%\bibliographystyle{ieee}
%\bibliography{bibliography}

\addtocontents{toc}{\protect\newpage}

% Include a single paper
%\input{beitrag}

%\includePaperPDF{paper_pdf}{\AuthorsTOC}{\AffiliationsTOC}{\Title}

\end{document}